\pdfoutput=1

\documentclass[11pt]{article}

\usepackage{acl}

\usepackage{times}
\usepackage{latexsym}
\usepackage{amsmath,amssymb,amsfonts}
\usepackage{graphicx}
\usepackage{tabularx}
\usepackage{multirow}
\usepackage{longtable}
\usepackage{booktabs}
\usepackage{hyperref}
\usepackage{subfigure}
\usepackage{dblfloatfix}    
\makeatletter
\DeclareRobustCommand{\cev}[1]{%
  \mathpalette\do@cev{#1}%
}
\newcommand{\do@cev}[2]{%
  \fix@cev{#1}{+}%
  \reflectbox{$\m@th#1\vec{\reflectbox{$\fix@cev{#1}{-}\m@th#1#2\fix@cev{#1}{+}$}}$}%
  \fix@cev{#1}{-}%
}
\newcommand{\fix@cev}[2]{%
  \ifx#1\displaystyle
    \mkern#23mu
  \else
    \ifx#1\textstyle
      \mkern#23mu
    \else
      \ifx#1\scriptstyle
        \mkern#22mu
      \else
        \mkern#22mu
      \fi
    \fi
  \fi
}

\usepackage{pifont}
\newcommand{\cmark}{\ding{51}}%
\newcommand{\xmark}{\ding{55}}%
\usepackage{fontawesome}

\usepackage{array}
\newcolumntype{L}[1]{>{\raggedright\let\newline\\\arraybackslash\hspace{0pt}}m{#1}}
\newcolumntype{C}[1]{>{\centering\let\newline\\\arraybackslash\hspace{0pt}}m{#1}}
\newcolumntype{R}[1]{>{\raggedleft\let\newline\\\arraybackslash\hspace{0pt}}m{#1}}

\usepackage[T1]{fontenc}

\usepackage[utf8]{inputenc}
\usepackage{subfiles}
\usepackage{microtype}

\usepackage{inconsolata}

\title{Align before Attend: Aligning Visual and Textual Features for Multimodal Hateful Content Detection}

\author{Eftekhar Hossain{\textsuperscript{\faDollar}}$^*$, Omar Sharif$^\mathbf{{\faPagelines}}${\textsuperscript{\faCny}}$^*$, Mohammed Moshiul Hoque{\textsuperscript{\faCny}}, Sarah M. Preum$^\mathbf{{\faPagelines}}$\\
   {\textsuperscript{\faCny}}Department of Computer Science and Engineering\\ 
  {$^{\faPagelines}$}Department of Computer Science, Dartmouth College, USA \\
\textsuperscript{{\faDollar}}Department of Electronics and Telecommunication Engineering\\ 
  {\textsuperscript{\faDollar \faCny}}Chittagong University of Engineering \& Technology, Bangladesh \\
 \texttt{\small \{eftekhar.hossain, moshiul\_240\}@cuet.ac.bd}, \texttt{\small\{omar.sharif.gr, sarah.masud.preum\}@dartmouth.edu}}

\begin{document}
\maketitle

\def\thefootnote{*}\footnotetext{Denotes equal contribution}

\begin{abstract}
Multimodal hateful content detection is a challenging task that requires complex reasoning across visual and textual modalities. Therefore, creating a meaningful multimodal representation that effectively captures the interplay between visual and textual features through intermediate fusion is critical. Conventional fusion techniques are unable to attend to the modality-specific features effectively. Moreover, most studies exclusively concentrated on English and overlooked other low-resource languages. This paper proposes a context-aware attention framework for multimodal hateful content detection and assesses it for both English and non-English languages. The proposed approach incorporates an attention layer to meaningfully align the visual and textual features. This alignment enables selective focus on modality-specific features before fusing them. We evaluate the proposed approach on two benchmark hateful meme datasets, \textit{viz.} MUTE (Bengali code-mixed) and MultiOFF (English). Evaluation results demonstrate our proposed approach's effectiveness with F1-scores of $69.7$\% and $70.3$\% for the MUTE and MultiOFF datasets. The scores show approximately $2.5$\% and $3.2$\% performance improvement over the state-of-the-art systems on these datasets. Our implementation is available at \href{https://github.com/eftekhar-hossain/Bengali-Hateful-Memes}{\textit{\textcolor{magenta}{https://github.com/eftekhar-hossain/Bengali-Hateful-Memes}}}.  
\end{abstract}
\textcolor{red}{\textbf{Disclaimer:} This paper contains hateful images that may be disturbing to some readers.}

\section{Introduction}
Recently, online platforms are witnessing an emerging trend of propagating hateful and offensive content. While most research in this area has focused on detecting hate speech from text-based content \citep{waseem2016hateful,schmidt-wiegand-2017-survey}, offensive multimodal content is also propagated, such as memes. Memes are images or screenshots with short texts embedded in them. Their sarcastic nature made them an increasingly popular tool for spreading hate and targeting individuals or communities based on various factors such as gender, race, ethnicity, religion, physical appearance, and sexual orientation \citep{WILLIAMS2016424,chhabra2023literature}. The proliferation of such content poses a significant threat to communal harmony and social stability and has therefore become an area of active research interest \citep{cao-etal-2022-prompting,pramanick-etal-2021-momenta-multimodal}.

\begin{figure}[t!]
  \centering
  \subfigure{\includegraphics[width =0.45\linewidth]{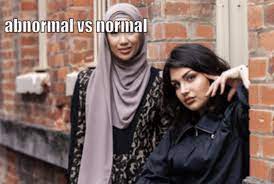}
  \label{intro-1}
  }\quad
  \subfigure{\includegraphics[width =.46\linewidth, height=2.4cm]{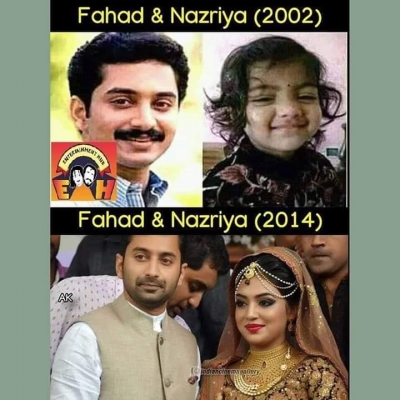}
  \label{intro-2}} 
\caption{Example of hateful memes. In isolation, neither the image nor the caption may appear hateful, but when combined, they can convey a hateful message.} 
 \label{intro-exmp}
\end{figure}

Multimodal hateful content detection requires a holistic understanding of visual and textual information. When considered separately, the image and caption components in Figure \ref{intro-1} may seem innocuous. The image portrays two women—one wearing a hijab and the other without and the caption states, ``abnormal and normal''. However, as a meme, this composition can be seen as derogatory towards the woman wearing the hijab by labeling her as abnormal. Similarly, the meme in Figure \ref{intro-2} insults the marriage of two South Indian celebrities by indicating their age gap in the text. Thus, focusing only on the image or the text is inadequate for complete understanding. Sometimes without the background information of the people and events used in a meme, it is difficult to interpret the meaning because the captions are short, fragmented, and sarcastic. Studies have demonstrated that off-the-shelf multimodal systems, which are typically effective in performing various visual-linguistic tasks, encounter difficulties when it comes to detecting hateful memes \cite{kiela2020hateful, cao-etal-2022-prompting}. Furthermore, the current state-of-the-art systems \cite{lee2021disentangling, pramanick-etal-2021-momenta-multimodal} for detecting hateful memes face limitations when applied to resource-constrained languages. This is primarily because several key components within their architectures are not accessible or well-supported in other languages. These challenges underscore the need for language-specific adaptations to address hateful meme detection in a broader linguistic context effectively.
    
To address this knowledge gap, we present a solution for detecting multimodal hateful memes. The approach leverages an attention-based context-aware fusion framework to create coherent multimodal representations. We hypothesize that by aligning visual and textual features before fusion, the network can better capture essential cues for accurate classification. The key challenge lies in effectively incorporating modality information to enable the network to focus on crucial features. Previous methods \cite{pramanick-etal-2021-momenta-multimodal, lee2021disentangling} used background context and additional captions while performing the fusion. In contrast, our approach introduces an attention layer to align modalities which simultaneously facilitates the extraction of contextual representations from both visual and textual modalities. Moreover, without adding external knowledge, the model's learning capability is augmented when the aligned representations are combined with modality-specific (i.e., visual, textual) features. To evaluate our approach, we conducted experiments on two benchmark datasets in different languages: MUTE \citep{hossain-etal-2022-mute} and MultiOFF \citep{suryawanshi-etal-2020-multimodal}. The evaluation results and ablation study demonstrate the effectiveness of our solution over baseline and state-of-the-art methods.

\noindent
The major contributions of this paper are three-fold:  \textit{(i)} develop an attention framework that effectively attends the contributing features of visual and textual modalities to detect multimodal hateful memes (Section \ref{p_framework}); \textit{(ii)} conduct an extensive evaluation on two different benchmark datasets on real-world memes to demonstrate the effectiveness of the proposed solution (Section \ref{e_results}, \ref{e_comparison}); and \textit{(iii)} perform ablation studies in different settings to examine the impact of BERT-base embeddings in detecting hateful memes while also investigate the model's quantitative and qualitative errors to understand its limitations (Section \ref{ablation}, \ref{error-a}).

\section{Related Work}

\textbf{Hateful Content Detection:} Over the past few years, offensive/hate speech detection has received a significant amount of attention from researchers. Some works focused on developing new corpus for different languages \citep{lekea2018detecting,roy2022hate} while others studied to develop novel methods \citep{li2022anti,mozafari2020bert}. However, most of the studies focused on hateful content detection from textual data and overlooked the multimodal aspects of the user-generated data. One such multimodal data is a meme, which combines both images and text.  With the flourishing of internet memes and because of their detrimental impact on society, online hateful meme classification got a considerable amount of traction from the research community \citep{das2020detecting, cao-etal-2022-prompting} lately. \newcite{suryawanshi-etal-2020-multimodal} and \newcite{kiela2020hateful} introduced hateful memes dataset in English. Besides developing datasets in English, few works attempted to introduce hateful memes datasets for low-resource languages such as 
Bengali \cite{hossain-etal-2022-mute}. 

\noindent
\textbf{Multimodal Fusion:} Over the years, various techniques have been applied to detect multimodal hateful memes. Conventional fusion \citep{vijayaraghavan2021interpretable,gomez2020exploring} by concatenating the modality-specific information is the most commonly used method for learning multimodal representation. Some works employed bilinear pooling \citep{chandra2021subverting} while others fine-tuned transformers \citep{kiela2020hateful} based architectures such as ViLBERT, MMBT, and Visual-BERT. Besides, some works attempted to use disentangled learning \citep{lee2021disentangling} and incorporate image captioning \citep{zhou2021multimodal} to improve the hateful memes detection performance. Recently, \newcite{cao-etal-2022-prompting} applied prompting techniques for hateful meme detection in English.  
To the best of our knowledge, no one has attempted to align the visual and textual features for hateful meme detection. Nonetheless, feature alignment is key in creating a successful multimodal representation \cite{pmlr-v162-zeng22c,NEURIPS2019_9fe77ac7}. This work aims to address this research gap by introducing an alignment technique for hateful meme detection.

Overall our work differs from the existing studies in several significant ways: \textit{(i)} rather than using additional context with conventional (i.e., early, late) fusion for multimodal representation, we align the visual and textual features using attention before fusing them, \textit{(ii)} Existing models are primarily designed for English and challenging to adapt for languages like Bengali. This work presents a model that uses alignment and can be adapted for any language by swapping out language-specific components, and \textit{(iii)} evaluation is performed on real-world meme dataset \textit{(MUTE, MultiOFF)} rather than the synthetic memes as in \newcite{kiela2020hateful}. 
\begin{figure*}[h!]
  \centering
  \includegraphics[width =0.95\linewidth]{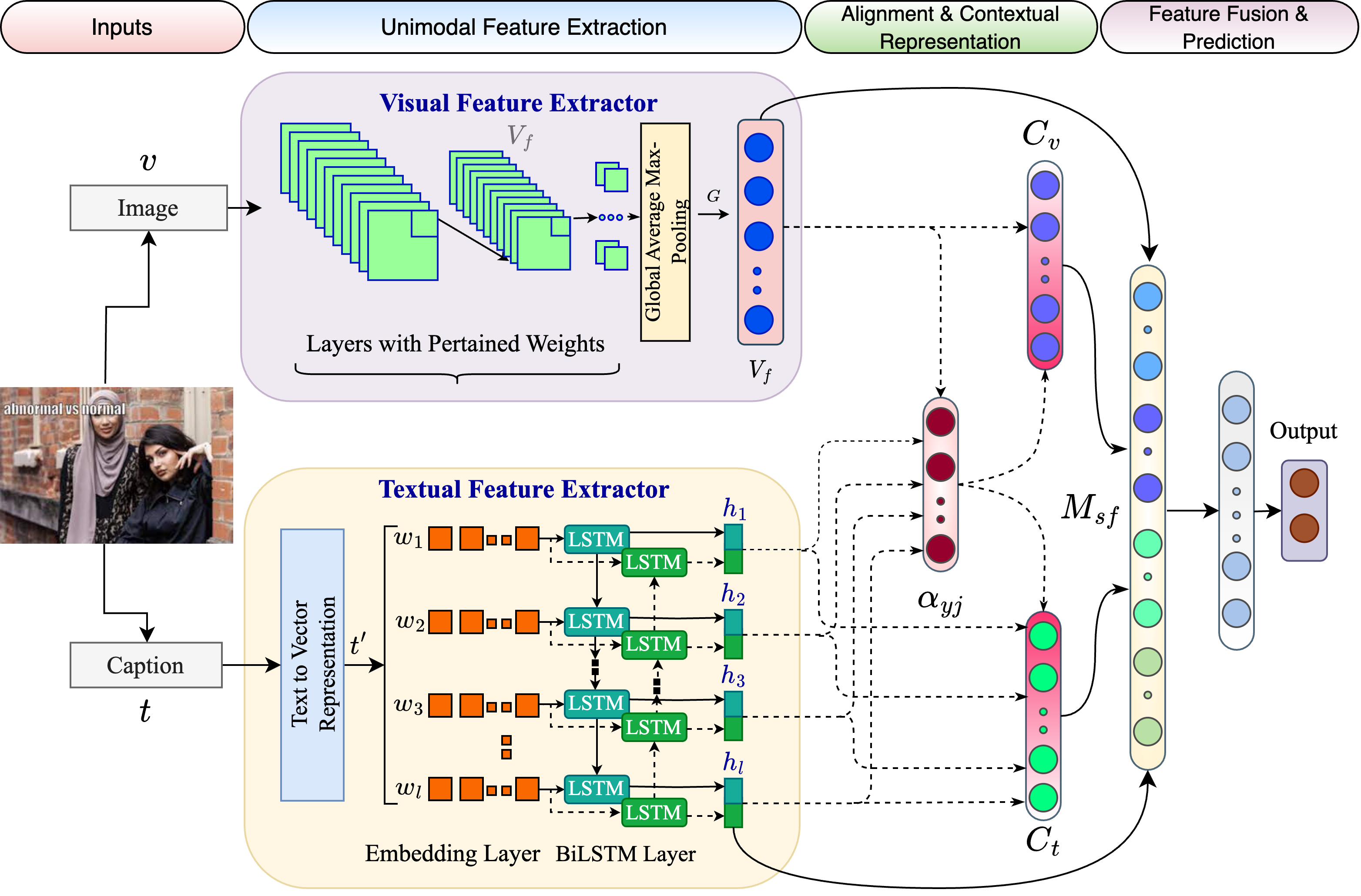}
 \caption{Our proposed context-aware multimodal architecture: $v$ and $t$ are the processed image and its corresponding caption. The upper block represents the visual feature extractor, and the lower block is the textual feature extractor. Alignment scores ($\alpha_{yj}$) are calculated by applying attention on visual ($V_f$) and textual ($h_1...h_l$) features. Subsequently, visual ($C_v$) and textual ($C_t$) context vectors are created by aligning ($V_f$) and ($h_1...h_l$) through alignment vector ($\alpha_{yj}$). Finally, by concatenating these context vectors ($C_v, C_t$) with modality-specific features ($V_f$, $h_l$) our method creates the multimodal context-aware representation $M_{sf}$.} 
 \label{block}
\end{figure*}

\section{Method}
Memes comprise two modalities (i.e., visual and textual); logically, one modality's content can outweigh another's content during prediction. Besides, not all the information from both modalities has an equal effect on determining whether a meme is hateful. We propose a context-aware fusion framework that selectively focuses on modality-specific information to model this complex relationship. The proposed network takes multimodal input and feeds the visual information to a CNN and textual information to an RNN for feature extraction. Then we calculate alignment weights over the visual and textual features through the attention layer. The objective is to capture the contributing features with higher weights by emphasizing both modalities. Subsequently, these alignment weights are utilized to create multimodal contextual representation. Finally, the resulting contextual and modality-specific representations are combined and passed to the softmax layer for classification. We denote our proposed architecture as \textbf{\textit{Multimodal Context Aware - Skip Connected Fusion (MCA-SCF)}} framework. An overall architecture of the framework is presented in Figure \ref{block}.

To ensure the robustness of the architecture, we experiment with three other variants of the proposed MCA-SCF framework: a) \textit{Vision Guided Contextual Fusion (VGCF)} framework; b) \textit{Text Guided Contextual Fusion (TGCF)} framework; and c) \textit{Multimodal Contextual Fusion (MCF)} framework. The architecture of these variants differs in context vector computation and information fusion. In \textit{VGCF}, we compute contextual information concerning the visual information and fuse it with the textual features. On the other hand, in \textit{TGCF}, the contextual information is computed with respect to textual features and integrated with the visual features. In contrast, we compute the context for both modalities and then combine them in \textit{MCF}. The rest of the components for all the architectures remain the same. The details of the variants \textit{VGCF, TGCF, MCF} can be found in Appendix \ref{app-variants}.

\subsection{Proposed (\textit{MCA-SCF}) Architecture}
\label{p_framework}
The \textit{MCA-SCF} framework consists of several components described in the following subsections. 

\subsubsection{Preprocessing}
Before feeding the data into the framework, we preprocess the visual ($v$) and textual ($t$) modality. For $v$, we resize the images to $150\times150\times3$ and transform the pixel values between $0$ to $1$ to reduce the computational complexity. On the other hand, we remove unwanted characters (i.e., symbols, URLs, numbers, etc.) from textual data. Then we encode each word with a unique number and make all the text lengths equal to size $l$, where $l$=$60$. 

\subsubsection{Visual and Textual Feature Extractor}
\label{fe}
We employ a pre-trained CNN (ResNet50) to obtain the visual features from the memes. We use ResNet50 because of its capability to address the vanishing gradient problem and effectiveness in several multimodal classification tasks \citep{hossain2022deep, hossain-etal-2022-memosen}. To adjust ResNet50, we exclude the top two layers from the main architectures and utilize the weights of the higher-level features previously trained on the ImageNet \citep{deng2009imagenet} dataset. We add a global average pooling layer followed by a dense layer and retrain the architecture with new weights. The following equation computes the visual features.
\begin{equation}
    V_{f} = Relu \left ( \sum_{k}^{d} W_{jk}*G_{k} + b_j \right )
\label{eqn:1}    
\end{equation}

Here, $V_{f}\in\mathbb{R}^{1 \times d}$ represents the visual semantic features extracted by the ResNet50 for the $m^{th}$ memes visual modality ($v$). Here, $d$ represents the number of neurons (100) in the dense layer. And, $G$ represents the feature map generated by the global average pooling layer while $W$ and $b$ represent the weight matrix and bias respectively.

We employ Recurrent Neural Network to extract both word-level and sentence-level textual features. Specifically, we use Bidirectional Long Short Term Memory (BiLSTM) network to capture the contextual dependency of the words. Initially, we generate the embedding vectors that give a semantic meaning to each word. The embedding dimension size is set to ($64$). The embedding vectors are passed to a BiLSTM which can keep the contextual dependency of the word vectors of $t$. The output of the BiLSTM network is generated by concatenating the forward and backward LSTM cell's output. It gives a word-level feature vector for every $k^{th}$ time step. The final time step ($l^{th}$) output is the sentence-level feature vector that we will use during the fusion operation. The features are computed using the following equation.

\begin{equation}
    h_j^{[k]} = \vec {h_j} \oplus \cev{h_j}
\label{eqn:3}    
\end{equation}
Here, $h_j^{[k]} \in \mathbb{R}^{1\times 2N}$ and $h^{[l]} \in \mathbb{R}^{1\times 2N}$ respectively denote the BiLSTM word-level and sentence level feature generated for $j^{th}$ word in the $k^{th}$ layer or time step. $l$ is padding length and $N$ is the number of hidden units (50) in the LSTM cell. The $\oplus$ represents the concatenation. All the hyper-parameter values are selected via trial and error fashion by monitoring the validation accuracy.

\subsubsection{ Alignment and Fusion }
Unlike existing approaches that employ early or late fusion techniques for multimodal representation, we align the visual and textual features through attention before joining them. Inspired from \cite{xu2015show} we apply the additive attention \citep{bahdanau2014neural} mechanism to develop the alignment model. The model assigns a score $\alpha_{y,j}$ to the world-level feature of the $j^{th}$ time step and the visual feature, $V_f$. The set of weights ${\alpha_{y,j}}$ determines how much image and text level features are aligned to predict a particular class label ($y$). The alignment score, $\alpha$ is parameterized by a feed-forward network where each feature vector (i.e., visual and textual) is trained with separate weights. The score function is therefore in the following form, given that $tanh$ is used as the non-linear activation function:  

\begin{equation}
\alpha (V_f, h_j) = v_{a}^T tanh \left(W_1*V_f + W_2*h_{j} \right) 
\label{eqn:4}
\end{equation}
\begin{equation}
\alpha_{y, j} = \frac{exp\left( \alpha (V_f, h_j)  \right)}{\sum_{j=1}^l exp\left( \alpha (V_f, h_j) \right) }
\label{eqn:5}
\end{equation}

After performing the softmax operation \eqref{eqn:5}, we obtain the normalized alignment scores, where higher weights are assigned to the feature combinations that are important for the prediction ($y$). Here, $v_a$, $W_1$, and $W_2$ are the weight matrices to be learned by the alignment model.

Afterward, we use these alignment scores to generate context vectors for each modality. The intuition behind this is that not all the features of individual modality are equally important for classification. Thus, focusing only on the significant feature is the key to better prediction. The following equation is computed for the context vectors.   

\begin{equation}
C_v = \sum_{j} \alpha_{y,j}* V_f
\label{eqn:6}
\end{equation}

\begin{equation}
C_t = \sum_{j} \alpha_{y,j}* h_j
\label{eqn:7}
\end{equation}

Here, $C_v \in \mathbb{R}^{1\times d}$ and $C_t \in \mathbb{R}^{1\times d}$ are referred to as the vision-guided and text-guided context vectors, respectively. These vectors keep the contextual and significant modality-specific information concerning both visual and textual modalities.

The context vectors are concatenated to generate a context-aware multimodal representation. Furthermore, inspired by the residual learning \citep{he2016deep} we concatenated each modality feature along with this contextual representation. The idea is to boost the gradient flow to the lower layer and enhance the multimodal representation. The following equation can express the combined feature representation.

\begin{equation}
M_{sf}  = C_v \oplus C_t \oplus V_f \oplus h^{[l]}
\label{eqn:8}
\end{equation}

Here, $M_{sf} \in \mathbb{R}^{1\times 4d}$ represents the contextual multimodal representation. This combined feature vector is then passed for the classification.

\section{Experiments and Results}
In this section, we first describe the datasets and the evaluation settings. We discuss the baselines and their results in comparison with the proposed method. Moreover, we conduct an ablation study to show how replacing components of the \textbf{\textit{MCA-SCF}} framework affects the performance. Subsequently, an error analysis will be provided to understand the model's error. Furthermore, we perform a cross-domain analysis to see how the proposed framework performs irrespective of language variation in a zero-shot setting (Appendix \ref{e_cross-domain}).  
\subsection{Datasets}
We train and evaluate our proposed approach on two benchmark multimodal datasets: the Multimodal Bengali Hateful Memes (MUTE) and a popular English Memes (MultiOFF) dataset. Due to the unavailability of datasets, we limited our performance assessment on these datasets. For this work we only consider real-world memes and avoid synthetic datasets \cite{kiela2020hateful}. Table \ref{tab:1} presents the distribution of the datasets.

\begin{table}[h!]
\centering
\small
\begin{tabular}{lc|cccc}
\textbf{Dataset}&\textbf{Class}&\textbf{Train}&\textbf{Validation}&\textbf{Test}\\
\hline

\multirow{2}{*}{\textbf{MUTE}}  & Hate & 1275 & 152 & 159 \\
& Not-Hate & 2092 & 223 & 257 \\

\midrule          
\multirow{2}{*}{\textbf{MultiOFF}}  & Offense& 187 & 59 & 59 \\
& Not-Offense &  258 & 90 &	90 \\
\hline

\end{tabular}
\caption{\label{tab:1} Distribution of MUTE and MultiOFF datasets. 
}
\end{table}

\begin{table*}[b!]
\centering

\scriptsize
\begin{tabular}{ll|cccc|cccc}

\hline
\textbf{Approach}&\textbf{Models} & \multicolumn{4}{c}{\textbf{MUTE}}& \multicolumn{4}{c}{\textbf{MultiOFF}}\\
\hline
&&\textbf{P}&\textbf{R}&\textbf{WF}&\textbf{AUC}& \textbf{P}& \textbf{R}&\textbf{WF}&\textbf{AUC}\\
\hline                   
\multirow{7}{*}{\textbf{Unimodal}}
 & ResNet50 (FT) & 0.634& 0.646 & $0.631_{\pm0.00}$ & $0.598_{\pm0.01}$ & 0.624 & 0.637 & $0.623_{\pm0.02}$ & $0.593_{\pm 0.01} $\\
 & ResNet50 (RT) &  0.617& 0.634 & $0.614_{\pm0.02}$ &	$0.580_{\pm0.03}$ & 0.580 & 0.557 &	$0.562_{\pm0.08}$ & $0.559_{\pm 0.01}$ \\
  & ViT &  0.622 & 0.639 & $0.584_{\pm0.03}$ &	$0.557_{\pm0.02}$ & 0.603 & 0.624 &	$0.559_{\pm0.06}$ & $0.542_{\pm 0.02}$ \\
 & BiLSTM & 0.660 & 0.670 & $0.658_{\pm0.02}$ & $0.626_{\pm0.02}$ & 0.611 & 0.604 &	$0.606_{\pm 0.02}$ &	$0.591_{\pm0.01}$ \\
 & BiLSTM + Attention & 0.659 & 0.622 & $0.627_{\pm0.02}$ & $0.636_{\pm0.01}$ & 0.577 & 0.597 & $0.578_{\pm0.02}$ & $0.548_{\pm0.01}$ \\
 & BERT & 0.645 & 0.658 & $0.642_{\pm0.08}$ & $0.609_{\pm0.06}$ & 0.621 & 0.617 & $0.610_{\pm0.01}$ & $0.611_{\pm0.09}$ \\
 & m-BERT & 0.627 & 0.644 & $0.620_{\pm0.02}$ & $0.586_{\pm0.01}$ & 0.584 & 0.611 & $0.574_{\pm0.02}$ & $0.547_{\pm0.07}$ \\
 & XLM-R & 0.646 & 0.656 & $0.648_{\pm0.04}$ & $0.618_{\pm0.01}$ & 0.612 & 0.630 & $0.580_{\pm0.01}$ & $0.557_{\pm0.08}$ \\
\midrule          

\multirow{6}{*}{\textbf{Multimodal}}& Early Fusion & 0.634 & 0.649 & $0.607_{\pm0.02}$ & $0.575_{\pm0.01}$ & 0.646 & 0.657 & $0.645_{\pm0.02}$ & $0.616_{\pm 0.06}$ \\            
& Late Fusion & 0.619 &	0.634 & $0.619_{\pm0.02}$ & $0.586_{\pm0.00}$ & 0.738 & 0.657 & $0.568_{\pm0.01}$ & $0.563_{\pm0.07}$ \\ 
& Attentive Fusion & 0.660 & 0.637 & $0.642_{\pm0.00}$ & $0.641_{\pm0.02}$ & 0.610 & 0.624 & $0.538_{\pm0.03}$ & $0.532_{\pm 0.06}$ \\   
& VisualBERT COCO & 0.494 & 0.572 & $0.530_{\pm0.04}$ & $0.521_{\pm0.01}$ & 0.396 & 0.689 & $0.503_{\pm0.07}$ & $0.502_{\pm 0.05}$ \\ 
& CLIP & 0.643 & 0.641 & $0.560_{\pm0.06}$ & $0.545_{\pm0.07}$ & 0.646 & 0.651 & $0.601_{\pm0.05}$ & $0.576_{\pm 0.03}$ \\ 
& ALBEF & 0.679 & 0.667 & $0.668_{\pm0.06}$ & $0.677_{\pm0.02}$ & 0.612 & 0.617 & $0.613_{\pm0.04}$ & $0.610_{\pm 0.04}$ \\ 
\midrule
\multirow{4}{*}{\begin{tabular}[c]{@{}c@{}}\textbf{Proposed System}\\ \textbf{and Variants}\end{tabular}}
& VGCF & 0.671 & 0.677 & $0.671_{\pm0.02}$ & $0.644_{\pm0.02}$ & 0.651 & 0.624 & $0.628_{\pm0.03}$ & $0.632_{\pm0.04}$ \\  
& TGCF & 0.662 & 0.665 & $0.663_{\pm0.01}$ &  $0.641_{\pm0.01}$  & 0.667 & 0.651 & $0.655_{\pm0.01}$ & $0.651_{\pm0.01}$ \\
& MCF & 0.692 & 0.699& $0.689_{\pm0.02}$ &  $0.659_{\pm0.01}$  & 0.654 & 0.657 & $0.655_{\pm0.05}$ & $0.635_{\pm0.04}$ \\
& \textbf{MCA-SCF (Proposed)} & 0.696 & 0.696 & {$\mathbf{0.697_{\pm0.00}}$} & {${0.674_{\pm0.01}}$} & 0.702 & 0.704 & {$\mathbf{0.703_{\pm0.02}}$} & \textbf{$0.686_{\pm0.03}$} \\
\hline
\end{tabular}
\caption{\label{result:1} Performance comparison of unimodal and multimodal models on test set where P, R, WF, and AUC denote precision, recall, weighted F1-score, and area under the receiver operating characteristics curve respectively. VGCF, TGCF, and MCF are the variants of the proposed MCA-SCF approach. The FT and RT represent the fine-tunned and retrained version of ResNet50, respectively. The standard deviation $(\pm)$ with five different random seeds is also reported. For space constraints, the score is not shown for precision and recall.   
}
\end{table*}
\paragraph{MUTE \citep{hossain-etal-2022-mute}:} A hateful memes dataset for the Bangla language. It consists of 4158 memes where the captions are code-mixed (Bangla + English) in nature. Among 4158 memes, 1586 are hateful and the rest of them are not hateful. We use the exact train-test split adopted by the authors to compare with our proposed approach. 

\paragraph{MultiOFF \cite{suryawanshi-etal-2020-multimodal}:} The MultiOFF consists of a total of 743 memes collected based on the US presidential election. The authors labeled the memes into the \textit{offensive} category. However, these memes can be considered hateful since they substantially overlap with the hatred category and contain derogatory/abusive content targeted toward a group of people. The training, validation, and test set contain 445, 149, and 149 memes.

We adopt the evaluation metrics from the previous works in hateful meme classification \citep{lee2021disentangling}. The superiority of a model is determined based on the weighted F1-score. Besides, weighted precision, recall, and Area Under the Receiver Operating Characteristics (AUC) scores have been reported for comparison. 
The details of the experimental settings are discussed in Appendix \ref{es}.
\subsection{Baselines}
We develop several baselines considering the unimodal (i.e., image or text) and multimodal information. The baseline models are chosen based on the best-performing models on these datasets (\textit{MUTE, MultiOFF}) and popular techniques from the existing literature. The model's hyperparameters are chosen via a trial-and-error approach by monitoring the validation accuracy. The baseline architectures are described in the following subsections.

\subsubsection{Unimodal Models}
\label{um}
Initially, we implemented models considering only the visual modality. We use the \textbf{ResNet50} network where we fine-tuned and retrained it with new weights. The architecture configuration kept the same as described in Section \ref{fe}. Besides, we also fine-tuned the \textbf{Vision Transformer (ViT)} \cite{vit} architecture on both datasets. On the other hand, for textual modality, we employed several architectures including \textbf{BiLSTM} \citep{baruah2019abaruah}, \textbf{BiLSTM + Attention} \citep{altin2019lastus}, \textbf{BERT} \citep{mozafari2020hate}, and \textbf{XLM-R} \citep{ranasinghe2020multilingual}. In one architecture we use an LSTM cell with 32 hidden units. Subsequently, the attention mechanism is added with the LSTM in another architecture. We use the language-specific variation of the BERT (i.e., \textbf{Bangla BERT} \citep{Sagor_2020} and \textbf{English-BERT} \citep{devlin2018bert}), the multilingual BERT (\textbf{m-BERT}), and cross-lingual BERT (\textbf{XLM-R}) for our task. We freeze the weights of these architectures and retrain them with new weights by adding a dense layer of 100 neurons. The dense layer takes the sentence embeddings as input and makes a higher-level representation of the text. Finally, this representation is passed to the classification layer for prediction.   

\subsubsection{Multimodal Models}
To develop the models using multimodal information, we use the most popular fusion techniques including \textbf{Early Fusion} \citep{pranesh2020memesem}, \textbf{Late Fusion} \citep{hossain-etal-2022-memosen}, and \textbf{Attentive Fusion} \citep{sharma2022r2d2}. We select the best-performing unimodal models (ResNet50 and LSTM) for visual and textual feature extraction. 
\begin{itemize}
    \item For early fusion, a dense layer of 100 neurons is added at both ends of individual modalities to make a joint representation by concatenating them.
    \item In late fusion, the classification layer's output from each modality is combined and then passed for the classification.
    \item With attentive fusion, the last dense layer's output is passed to an attention layer, and then the resulting attentive vector is used for classification. 
\end{itemize} 
Finally, we employed several state-of-the-art multimodal architectures including \textbf{VisualBERT-COCO} \cite{li2019visualbert}, \textbf{CLIP} \cite{radford2021learning}, and \textbf{ALBEF} \cite{li2021align} and fine-tuned them on our datasets.

\subsection{Results}
\label{e_results}
Table \ref{result:1} presented the outcome of the baselines and proposed method over the test set of \textit{MUTE} and \textit{MultiOFF} datasets. In \textit{MUTE}, the visual models (ResNet50 and ViT) failed to obtain a satisfactory outcome, while among the textual models, BiLSTM achieved the highest F1-score of 0.658. Surprisingly, the performance of the pre-trained transformers is lower than BiLSTM. We perform a detailed ablation study to get more insights on this. Meanwhile, when multimodal information is integrated, the attentive fusion approach achieved the highest F1 (0.642) and AUC (0.641) scores compared to its counterparts (i.e., early and late fusion). Among the other multimodal architectures (i.e., VisualBERT, CLIP, and ALBEF), ALBEF showed outstanding performance with an F1 score of 0.668. However, we observed that the variants (VGCF, TGCF, and MCF) of the alignment approach obtained superior performance over the unimodal and other multimodal models except ALBEF. Even though they achieved better outcomes, the proposed MCA-SCF framework outperformed all the models by getting the highest F1 score of 0.697.

In \textit{MultiOFF} dataset, BERT achieved the highest F1-score of 0.610 amid the unimodal models. On the other hand, early fusion showed significantly higher performance (0.645) compared to late fusion (0.568), attentive fusion  (0.538), and other multimodal architectures such as VisualBERT (0.503), CLIP (0.601), and ALBEF (0.613). We noticed that the performance is substantially improved with the variants. Nonetheless, MCA-SCF outperforms all the models, obtaining the highest F1 score of 0.703 and AUC score of 0.686. 

In summary, the proposed \textit{MCA-SCF} framework and its variants outperformed the baselines in both datasets. Aligning the visual and textual information before fusing them played a crucial role in boosting the model's predictive performance.

\subsection{Ablation Study}
\label{ablation}
In addition to the experiments emphasizing the importance of context-aware multimodal representation for hateful meme classification in Table \ref{result:1}, we also examine the effect of contextualized embeddings in MCA-SCF instead of simple word embeddings. We consider three transformer models i.e., language-specific BERT, multilingual BERT, and XLM-R. We employed the architecture with similar parameters described in Section \ref{um}. Two individual models were developed for each transformer architecture. Firstly, BERT word level and sentence level embeddings were used to develop MCA-SCF whereas in the second case, contextualized embeddings were passed to an LSTM layer and utilized the LSTM word level features with the contextualized sentence embeddings to construct MCA-SCF. The training parameters of the models were kept the same as discussed in Appendix \ref{es}.
Table \ref{as} reported the outcomes when contextualized embeddings are used. We observed that, in the case of \textit{MUTE}, MCA-SCF with m-BERT obtained the highest F1 score (0.665), whereas MCA-SCF with m-BERT + BiLSTM achieved the maximum F1 score (0.613) in \textit{MultiOFF} dataset. 

\begin{table}[t!]
\centering
\scriptsize
\begin{tabular}{l|cc|cc}

\hline
\textbf{Models} & \multicolumn{2}{c}{\textbf{MUTE}}& \multicolumn{2}{c}{\textbf{MultiOFF}}\\
\hline
&\textbf{WF}& \textbf{AUC}& \textbf{WF} & \textbf{AUC}\\
\hline                  
 MCA-SCF w/ BERT + BiLSTM & 0.657 & 0.634 & 0.571 & 0.542  \\
 MCA-SCF w/ only BERT & 0.649 & 0.637  & 0.612 & 0.586 \\
 MCA-SCF w/ m-BERT + BiLSTM & 0.645 & 0.622 & 0.613 & 0.589 \\
 MCA-SCF w/ only m-BERT & 0.665 & 0.676 & 0.575 & 0.551 \\
 MCA-SCF w/ XLM-R + BiLSTM & 0.615 & 0.582 & 0.525 & 0.501 \\
 MCA-SCF w/ only XLM-R & 0.661 & 0.627 & 0.540 & 0.513 \\       
\hline
\end{tabular}
\caption{\label{as} Effect on the proposed method performance when replacing the text model with various transformer architectures.  
}
\end{table}

The findings reveal that there is no significant effect of using the BERT-based models for hateful meme detection. Even the BERT-based model outcomes are lower than the variants of the proposed method. Therefore, it can be stated that contextualized embeddings are not suitable for hateful meme detection. The reason behind this lower performance could be the fact that the memes' captions are very different from regular texts. BERT-based models are typically trained on longer and more complete textual inputs, whereas the language used in meme captions is often short, fragmented, and sarcastic. This discrepancy in language style can cause this suboptimal performance.

\begin{table}[b!]
\centering
\small
\begin{tabular}{ll|c}

\hline
\textbf{Dataset}&\textbf{Approaches}&\textbf{WF (\%)}\\
\hline                   
\multirow{2}{*}{\textbf{MUTE}}
 & \newcite{hossain-etal-2022-mute} & 67.2  \\
 & Proposed & \textbf{69.7}\\
\midrule          
\multirow{4}{*}{\textbf{MultOFF}} 
& \newcite{suryawanshi-etal-2020-multimodal} & 54.0  \\
& \newcite{lee2021disentangling} & 64.6 \\
& \newcite{hossain2022identification} & 66.7 \\
& \newcite{zhong2022combining}  & 67.1\\
& Proposed & \textbf{70.3}\\
\hline
\end{tabular}
\caption{\label{comparison} Comparative analysis of the proposed method with the existing state-of-the-art systems.} 
\end{table}

\label{error-a}

\begin{figure*}[t!]
  \centering
  \subfigure[MUTE]{\includegraphics[width =0.45\textwidth]{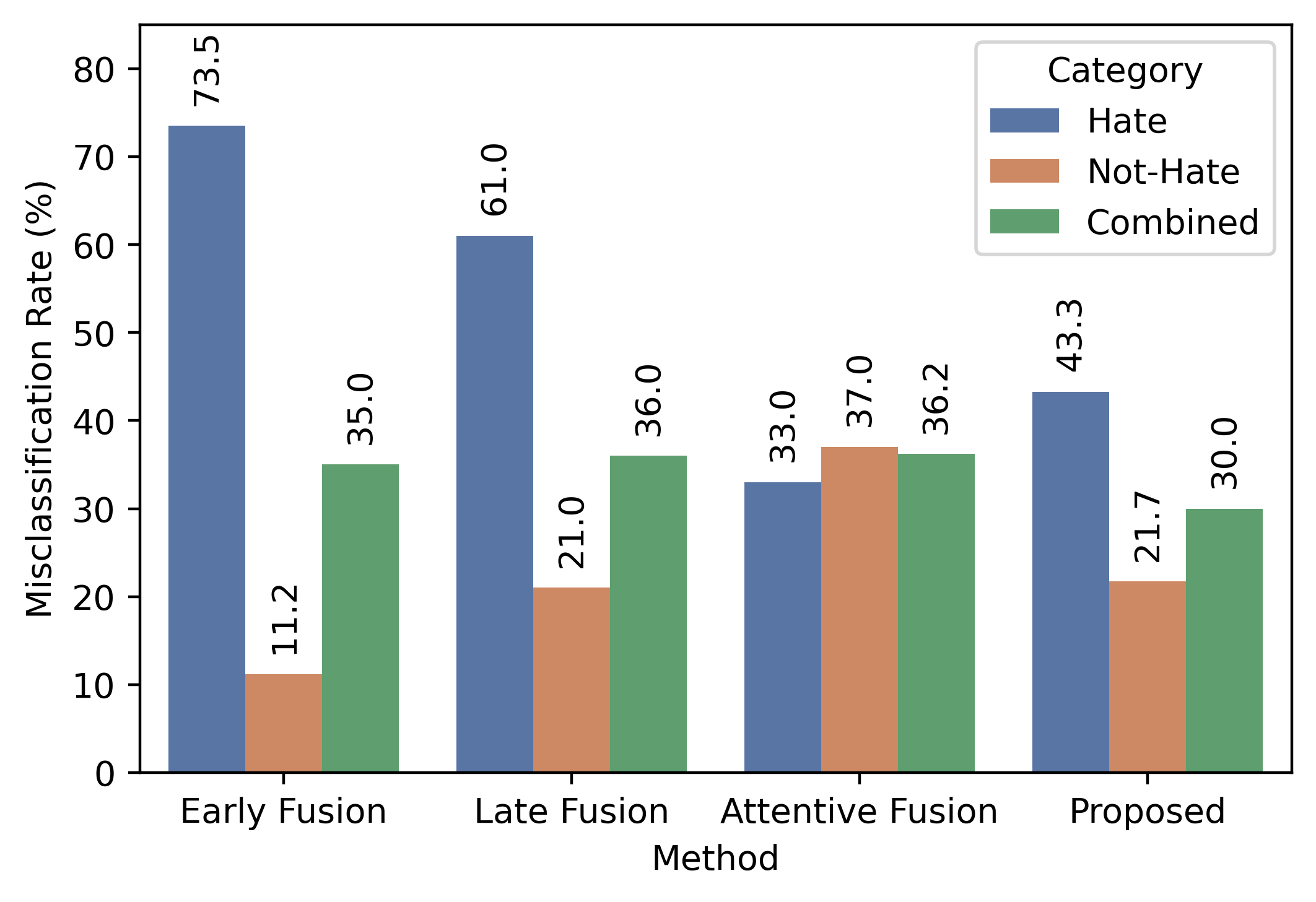}}\quad
  \subfigure[MultiOFF]{\includegraphics[width =0.45\textwidth]{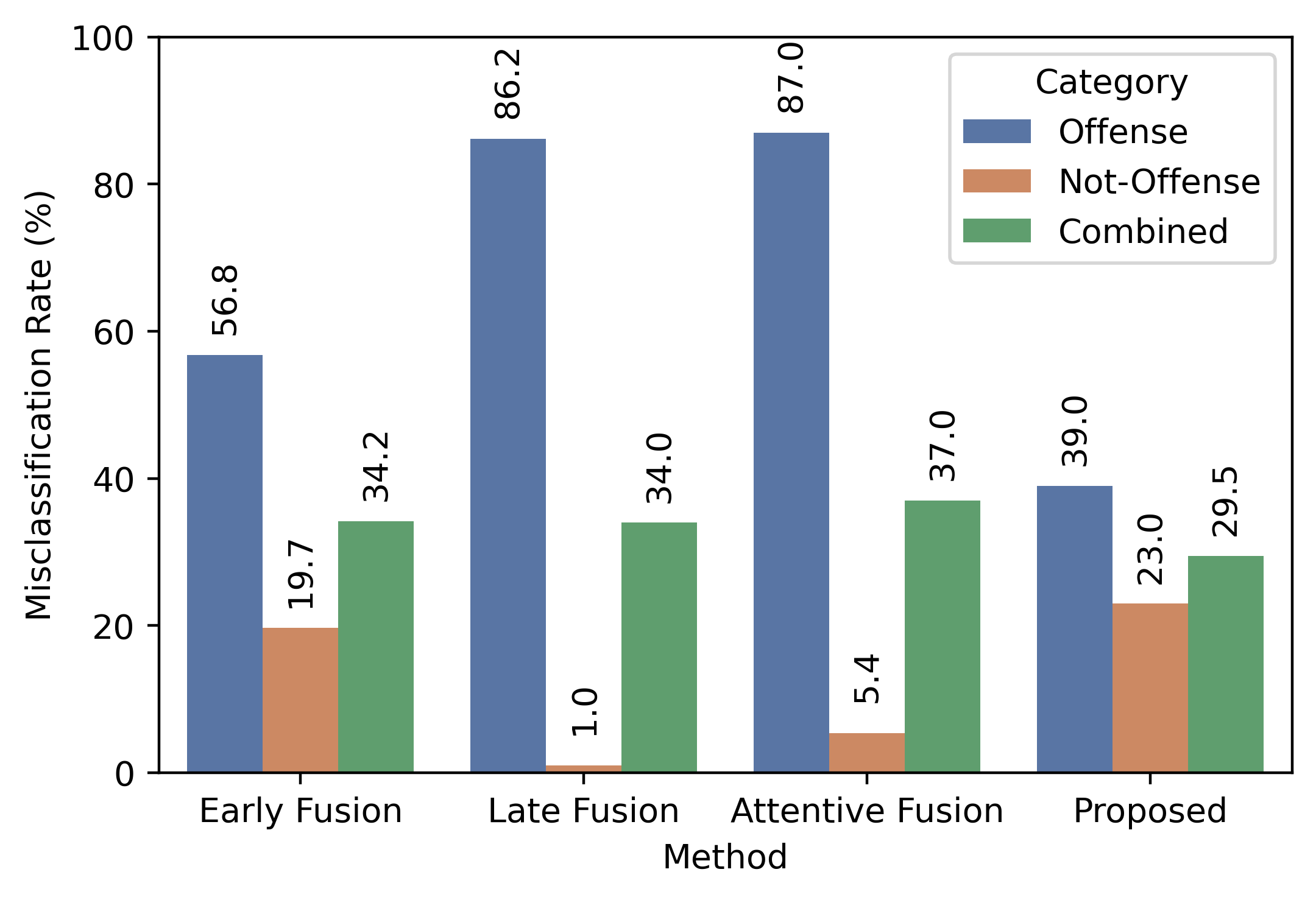}} 
 \caption{Misclassification rate comparison between various fusion approaches (i.e., early, late, attentive) and proposed (MCA-SCF) method on both datasets.} 
 \label{error}
\end{figure*}

\subsection{Comparison with Existing Studies}
\label{e_comparison}
Table \ref{comparison} presents the performance comparison of the proposed method with the existing state-of-the-art systems on the datasets. In \textit{MUTE}, our proposed multimodal framework achieves the best F1 score of 69.7\% ($\uparrow$ 2.5\%) as compared to the existing highest score of 67.2\%. Likewise, for \textit{MultiOFF} dataset, we obtain the highest F1 score of 70.3\% ($\uparrow$ 3.2\%) beating the current state-of-the-art system (67.1\%). The performance improvement in both datasets' indicates our proposed method's novelty.

\subsection{Error Analysis}
We investigate the errors of the proposed MCA-SCF approach both quantitatively and qualitatively.\\

\noindent
\textbf{Quantitative Analysis:} Early, late, and attentive fusion techniques have been considered to compare the errors with the proposed approach. We measured the Misclassification Rate (MR) for all the models reported in Figure \ref{error}. For \textit{MUTE} dataset, we observed that the MR is reduced at 43.3\% (proposed method) from 73.5\% (early fusion) in \textit{Hate} class while it is increased $\approx$10\% in \textit{Not-Hate} class. However, the error rate in \textit{Not-Hate} class is minimal with the early fusion approach, whereas for \textit{Hate} class, the attentive fusion approach reduces the error most. To conclude, we computed the combined class error rate and found that the overall system's error is the lowest (30\%) with the proposed MCA-SCF method. Likewise, in \textit{MultiOFF}, the proposed method achieves the lowest combined error rate of 29.5\%. It is worth noting that the proposed model significantly reduces the error rate in negative classes, enabling effective detection of hateful memes. One interesting aspect observed is that the misclassification rate is higher in the Negative (\textit{Hate or Offense}) class compared to the Positive (\textit{Not-Hate or Not-Offense}) class across all approaches. This discrepancy could be attributed to the uneven distribution of data, with fewer training samples in the negative classes. As a result, the model may have struggled to effectively learn visual and textual patterns, leading to incorrect predictions.

\begin{figure}[t!]
  \centering
  \subfigure[\textbf{EF:} Not-Hate (\xmark) \newline \textbf{AF:} Not-Hate (\xmark)  \newline \textbf{Proposed:} Hateful (\cmark)]{\includegraphics[scale=0.4,width =3.5cm,height = 2.5cm]{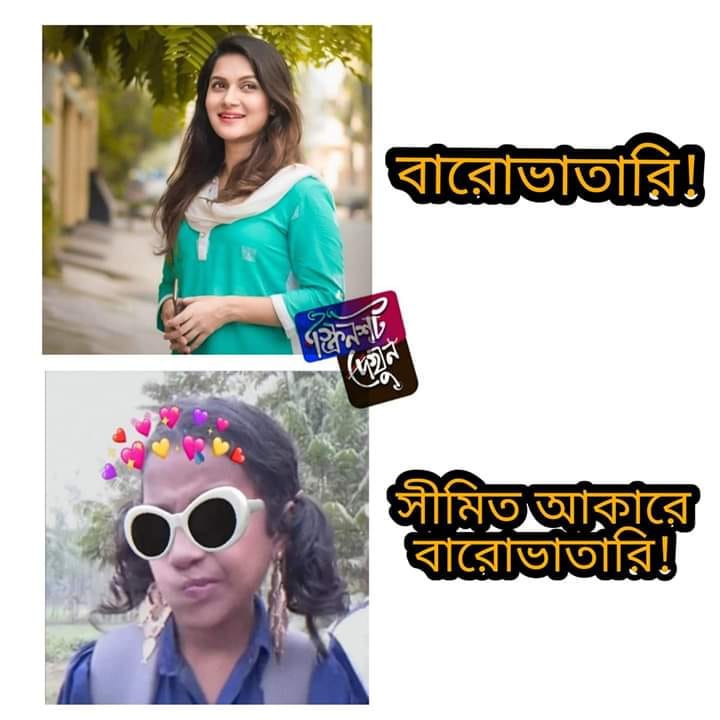}}\quad
  \subfigure[ \textbf{Actual:} Hateful  \newline  \textbf{Predicted:} \textcolor{red}{Not-Hate}]{\includegraphics[scale=0.4,width =3.5cm,height = 2.5cm]{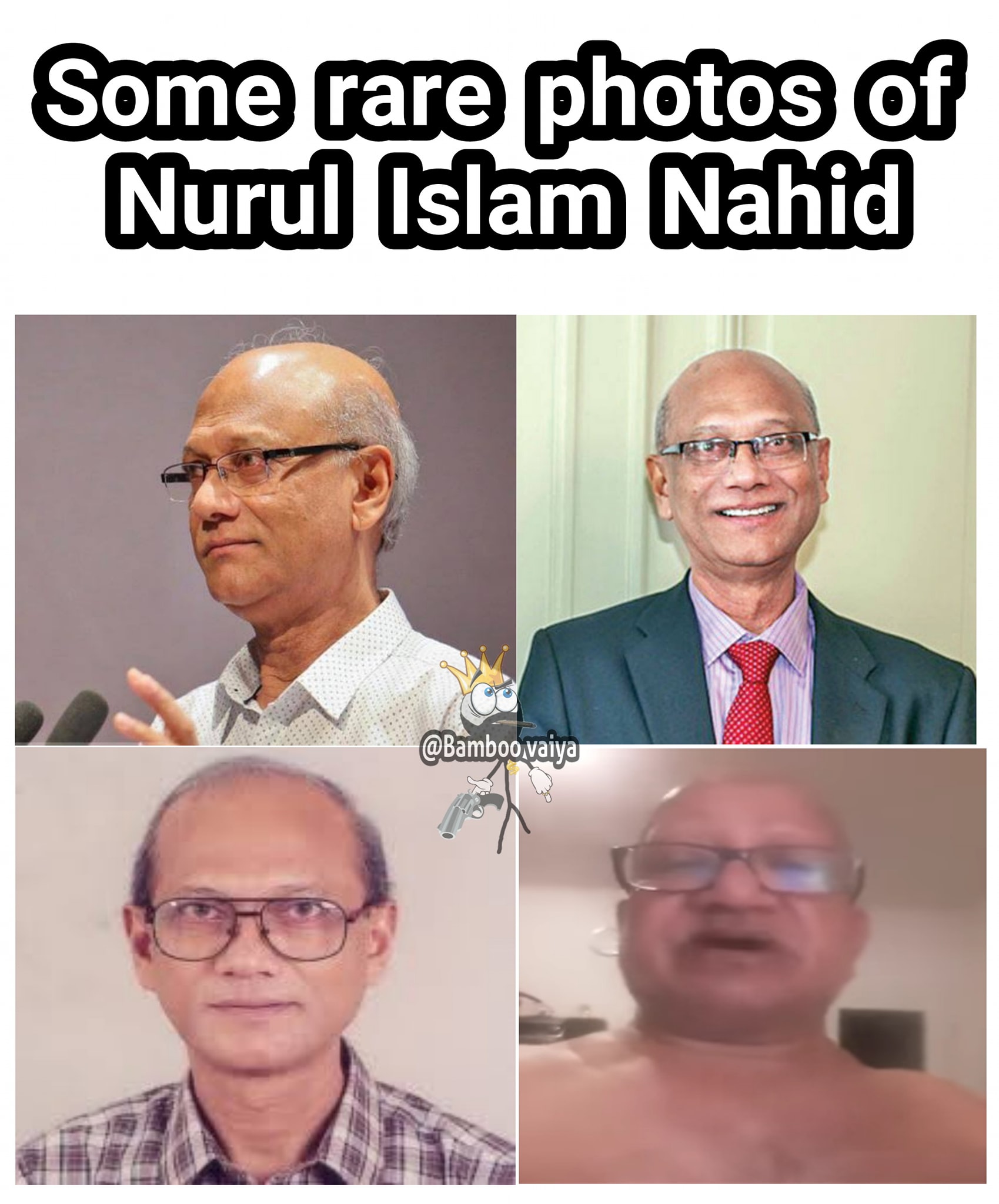}}

 \caption{Example (a) shows a meme where the proposed method yields better predictions, and example (b) illustrates a wrongly classified sample. The symbol (\cmark) and (\xmark) indicates the correct and incorrect prediction. EF and AF represent the early fusion and attentive fusion approaches, respectively.} 
 \label{qa}
\end{figure}

\noindent
\textbf{Qualitative Analysis:} We also perform qualitative analysis by investigating model predictions on a few samples.  For example, the meme in Figure \ref{qa} (a) was misclassified as \textit{Not-Hate} by the early and attentive fusion approaches. However, the proposed method captures the image and textual features that represent the context of the meme and therefore can correctly predict them as \textit{Hateful}. We also analyze where the proposed method failed to give accurate inferences. For instance, the model misclassified the meme shown in Figure \ref{qa} (b) as \textit{Not-Hate} when the actual label is \textit{Hate}. The reason for this misclassification could be the presence of consistent visual features ``Bald Man'' and the absence of any trigger word in the text. Moreover, the model needs world-level knowledge to understand that this meme is demeaning the identity of a reputed person in Bangladesh. The above analysis shows that we need to explore more advanced reasoning modules to classify such memes accurately.

\section{Conclusion}
This paper presents \textit{MCA-SCF}, a multimodal framework that aligns visual and textual features using attention to create a coherent contextual representation. The model aims to improve hateful content detection performance by leveraging contextual and modality-specific representations. We evaluate the model on two publicly available datasets i.e., \textit{MUTE} and \textit{MultiOFF}. Our extensive experiments demonstrate that \textit{MCA-SCF} outperforms the state-of-the-art systems on these datasets. Furthermore, we conducted experiments with different variants of the model and performed an ablation study to ensure the system's robustness. The ablation study reveals that general word embeddings are more suitable than contextualized embedding for multimodal hateful meme detection. Finally, the cross-domain analysis illustrates the model's generalization capability in zero-shot settings. 

\section*{Limitations}
We identify several findings in this work.  Firstly, we found that advanced multimodal models (e.g., CLIP, and VisualBERT) can not show satisfactory performance on both datasets. One compelling reason can be attributed that these models are not pretrained on enough Bengali image-text pairs and thus perform poorly when fine-tuning on the MUTE dataset. On the other hand, the lags in the performance in MultiOFF due to having fewer samples. As a result, the model does not get enough examples to learn complex relationships in the task and provides inferior performance. Besides that other advanced multimodal models (i.e., ALIGN, FLAVA, ViLBERT, BLIP) are rarely pretrained for Bengali image text pairs, limiting their applications in such low-resource languages. Therefore, we focus on enhancing the performance of off-the-shelf models with minimal computation by improving intermediate fusion through alignment. Our error analysis indicates that there is still significant room for improvement to effectively align visual and textual features for multimodal hateful content detection. Secondly, while the proposed model can infer the implicit meaning of memes in certain cases, it still falls short in complex reasoning to comprehend the contextual nuances of memes with concise captions. Finally, due to the unavailability of real-world meme datasets, we limited our performance assessment to two benchmark datasets. In the future, we plan to apply the model to detect memes in similar domains like harm and aggression, demonstrating its robustness across diverse and challenging categories.



\bibliography{eacl24}
\bibliographystyle{acl_natbib}

\appendix
\section*{Appendix}
\label{sec:appendix}
\counterwithin{figure}{section}
\counterwithin{table}{section}

\section{Variants of \textit{MCA-SCF} Framework }
\label{app-variants}
\begin{figure*}[t!]
  \centering
  \includegraphics[width =1\linewidth]{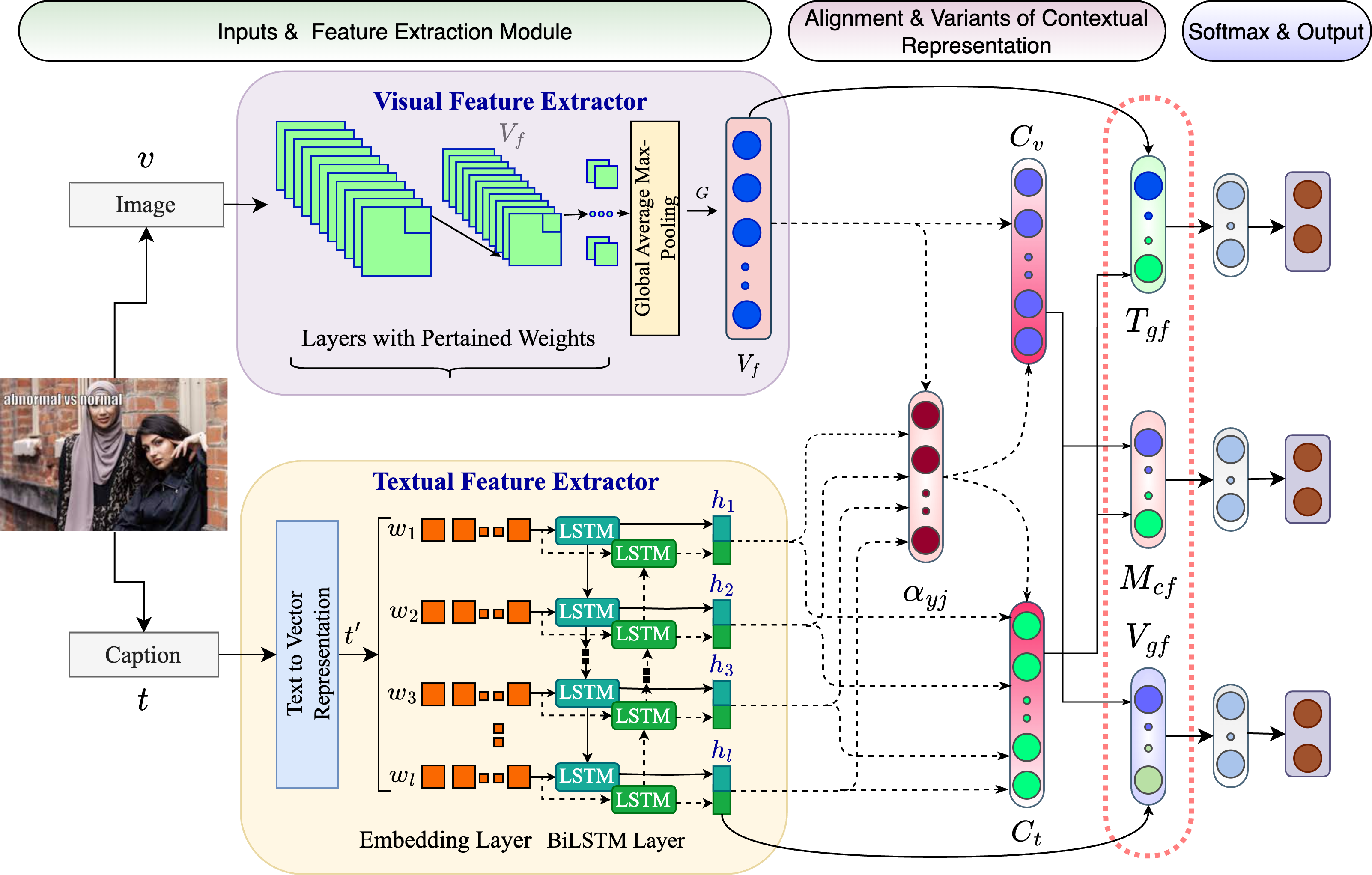}
 \caption{Variants of the proposed MCA-SCF framework. The majority of the components remain the same as illustrated in figure \ref{block}. The three variants ($V_{gf}, M_{cf}, T_{gf}$) have differences in the way they integrate information to emphasize the context of a particular modality. } 
 \label{variants}
\end{figure*}

We develop three other variants of the \textit{MCA-SCF} network namely \textit{VGCF}, \textit{TGCF}, and \textit{MCF}. Figure \ref{variants} shows the computation of the variants. The \textit{VGCF} framework does not account for the context vector generated from the text modality. After aligning the visual and textual modalities, we used the obtained alignment score ($\alpha_{yj}$) to highlight only the significant visual information and combined them with the sentence-level ($h^{[l]}$) textual feature. The VGC vector $V_{gf} \in \mathbb{R}^{1\times 2d}$ can be expressed by the following equation.

\begin{equation}
V_{gf}  = C_{v} \oplus h^{[l]}
\label{eqn:9}
\end{equation}

On the other hand, with \textit{TGCF} framework, we utilize the alignment score to generate a contextual representation ($C_t$) only for the text modality. This representation is then combined with the visual features ($V_f$) to compute the TGC vector $T_{gf} \in \mathbb{R}^{1\times 2d}$  by the equation \eqref{eqn:10}.

\begin{equation}
T_{gf}  = C_{t} \oplus V_f
\label{eqn:10}
\end{equation}

In the \textit{MCF} framework, we combined the two context vectors (i.e., $C_v$ and $C_t$) to make a contextual multimodal representation. The vector $M_{cf} \in \mathbb{R}^{1\times 2d}$ can be expressed by the equation.

\begin{equation}
M_{cf}  = C_{v} \oplus C_{t} 
\label{eqn:11}
\end{equation}
\section{Experimental Settings}
\label{es}
We perform experiments on the Google Colab platform. The transformer architectures were downloaded from the huggingface library and implemented using the TensorFlow framework. All the models are compiled using \textit{binary cross-entropy} loss function. For all the models the error optimization is performed by the \textit{Adam} optimizer with a learning rate of $1e^{-3}$ except for the transformer-based models which are $3e{-5}$. We used the \textit{batch size} of 16 and trained the models for 20 \textit{epochs}. To save the best intermediate models during training Keras checkpoint method has been utilized. 

\section{ Zero-shot Cross-Domain Transfer}
\label{e_cross-domain}

We examine the cross-domain transfer ability of the proposed method by training it on a source dataset and evaluating it on a target dataset. Besides, we also investigate the proposed method's performance when the training is done on combined datasets but tested only on a particular dataset. We focus on examining the impact of captured phenomena between datasets. The cross-domain performance has been measured by the relative zero-shot transfer ability \citep{turc2021revisiting}. We denoted it as the recovery ratio because it indicates the ratio of how much performance is recovered by changing the source domain and it is given as follows.

\begin{equation}
    R(S,T) = \frac{F(S,T)}{F(T,T)}
\end{equation}

Here, $F(S,T)$ is a model performance (i.e., $f_1$-score) for the source domain $S$ on the target domain $T$. If the source and target domains are the same, the $R$ would be $1.0$. The recovery scores of both zero-shot and combined dataset settings are given in Table \ref{cdt}.

\begin{table}[h!]
\centering

\small
\begin{tabular}{clcC{1cm} C{1cm}}

 &  & & \multicolumn{2}{c}{\textbf{Target}} \\
 \hline
 & & & MUTE & MultiOFF \\
 \hline

\multirow{7}{*}{\rotatebox{90}{\textbf{Source}}}&\multirow{3}{*}{{\textbf{Zero-shot}}} & MUTE & 0.697 & 0.585 (84\%) \\
\cline{3-5}
& & MultiOFF & 0.527 (75\%) & 0.703 \\
\cline{2-5}
 &\multirow{1}{*}{\textbf{Cross- domain}} & \multirow{1}{*}{MT+MO}& 0.604 (86\%) & 0.627 (90\%)  \\
 \hline
\end{tabular}

\caption{\label{cdt} Effect of the zero-shot and cross-domain transfer on both datasets. MT+MO indicates the combination of the MUTE and MultiOFF datasets. The major diagonal represents the actual performance, while the minor diagonal indicates how much performance is recovered when we change the source dataset.}
\end{table}

In both settings, the recovery rate is comparatively higher when we evaluate on \textit{MultiOFF} dataset and train using the \textit{MUTE} dataset. For instance, in the zero-shot setting, the \textit{MUTE} dataset $75\%$ performance of $0.697$ is recovered when the source domain was the \textit{MultiOFF} dataset. Similarly, we observed that $84\%$ is the recovery rate on \textit{MultiOFF} when \textit{MUTE} is the source domain. On the other hand, with a combined setting, $86\%$ and $90\%$ performance is recovered of the \textit{MUTE} and \textit{MultiOFF} datasets. Overall, in zero-shot setting \textit{MUTE} as a source dataset can mostly recover the performance from \textit{MultiOFF}. This could happen because \textit{MUTE} consists of code-mixed captions and has more training samples. This may allow for a greater transfer and sharing of multimodal features between datasets, ultimately contributing to the model's strong performance on the \textit{MultiOFF} dataset. Meanwhile, the proposed method can not generalize well on MUTE when trained with \textit{MultiOFF} dataset. This is because the less number of training samples in \textit{MultiOFF} and the model do not get any information about the Bengali language from the dataset. In contrast to its moderate generalization performance in the zero-shot setting, our proposed method demonstrates strong performance in the test set of each dataset when trained on the combined training set.

\end{document}